\documentclass[10pt,twocolumn,letterpaper]{article}

\usepackage[pagenumbers]{cvpr} 

\usepackage[dvipsnames]{xcolor}

\definecolor{cvprblue}{rgb}{0.21,0.49,0.74}
\usepackage[pagebackref,breaklinks,colorlinks,citecolor=cvprblue]{hyperref}

\usepackage{amsmath}
\usepackage{graphicx} 
\usepackage{multirow}
\usepackage{multicol}
\usepackage{bm}

\def\paperID{*****} 
\def\confName{CVPR}
\def\confYear{2024}

\title{Snakes and Ladders: Two Steps Up for VideoMamba}

\author{%
  Hui Lu\thanks{Corresponding author}, Albert Ali Salah,  Ronald Poppe\\
  Utrecht University\\
}
\begin{document}
\maketitle

\begin{abstract}
Video understanding requires the extraction of rich spatio-temporal representations, which transformer models achieve through self-attention. Unfortunately, self-attention poses a computational burden. In NLP, Mamba has surfaced as an efficient alternative for transformers. However, Mamba's successes do not trivially extend to vision tasks, including those in video analysis. In this paper, we theoretically analyze the differences between self-attention and Mamba. We identify two limitations in Mamba's token processing: historical decay and element contradiction.
We propose VideoMambaPro (VMP) that solves the identified limitations by adding masked backward computation and elemental residual connections to a VideoMamba backbone. Differently sized VideoMambaPro models surpass VideoMamba by 1.6-2.8\% and 1.1-1.9\% top-1 on Kinetics-400 and Something-Something V2, respectively. Even without extensive pre-training, our models present an increasingly attractive and efficient alternative to current transformer models. Moreover, our two solutions are orthogonal to recent advances in Vision Mamba models, and are likely to provide further improvements in future models.
\end{abstract}

\section{Introduction}
Video understanding is a challenging task, requiring models that can extract rich spatio-temporal representations from video inputs. Transformers are powerful neural networks capable of effectively capturing temporal and spatial information from videos~\cite{li2022uniformerv2,lu2024enhancing,wang2023videomae}. Therefore, most current state-of-the-art models for video understanding are based on transformers~\cite{ryali2023hiera,wang2022internvideo}.
%
At the core of transformers is self-attention~\cite{vaswani2017attention}, which learns the self-alignment between tokens in an input sequence by estimating the relative importance of a given token with respect to all other tokens. This long-range token dependency accounts for much of the success of transformer models~\cite{dosovitskiy2021an,vaswani2017attention}. 

The cost involved in computing self-attention is high, which eventually limits the application of powerful transformer models in practical settings~\cite{keles2023computational}. Recently, alternative models with lower-cost operators have been proposed for national language processing (NLP), including S4~\cite{gu2021efficiently}, RWKV~\cite{peng2023rwkv}, and RetNet~\cite{sun2023retentive}. Among these methods, Mamba~\cite{gu2023mamba} shows the best performance on long-range and causal tasks such as language understanding~\cite{mehta2022long} and content-based reasoning~\cite{patro2024mamba}.

Motivated by the favorable computational cost, researchers have recently extended Mamba from the NLP domain to the computer vision domain. The core adaptation involved splitting the input image into multiple regions and embedding these as continuous tokens \cite{zhu2024vision}. For video understanding, the recently proposed VideoMamba~\cite{li2024videomamba} extracts key frames from videos as the continuous input sequence. However, compared to previous transformer-based methods, VideoMamba's performance on video benchmarks is markedly lower. For example, VideoMamba achieves 82.4\% top-1 on Kinetics-400, compared to 85.2\% for VideoMAE~\cite{tong2022videomae}, indicating room for improvement.

In this paper, we first analyze differences in the feature extraction capabilities of transformers and Mamba. We identify two limitations of Mamba when applied to video understanding: historical decay and element contradiction. We then extend VideoMamba to mitigate these limitations. 
Note that there are different works named VideoMamba, and our work is based on~\cite{li2024videomamba}.The proposed VideoMambaPro (VMP) addresses historical decay through masked backward computation in the bi-directional Mamba process, allowing the network to better handle historical tokens. We introduce residual connections to Mamba's matrix elements to tackle element contradiction. VideoMambaPro consistently improves the performance of VideoMamba on video understanding tasks, positioning it as a strong, efficient competitor to transformers. Our contributions are:
\begin{itemize}
    \item We derive a formal representation of Mamba from the perspective of self-attention and identify two limitations of Mamba in the video analysis domain.
    \item We propose VideoMambaPro, effectively addressing Mamba's limitations for video understanding.
    \item We report strong video action recognition performance compared to recent Vision Mamba-based models, and surpass the original VideoMamba by clear margins.
\end{itemize}

We first discuss related work. Then, we provide our theoretical analysis, before introducing the VideoMambaPro architecture. Experiments are summarized in Section~\ref{sec:experiments} and we conclude in Section~\ref{sec:conclusion}

\section{Related Work}
\textbf{Transformers.} One core aspect of transformers is self-attention~\cite{vaswani2017attention} to achieve long-range interactions by measuring the similarity between tokens. Self-attention was introduced in the computer vision domain for tasks such as image recognition~\cite{liu2023simpleclick,strudel2021segmenter} and object detection~\cite{fang2021you,zhang2021vit}. Subsequent works (e.g., \cite{feichtenhofer2022masked,li2023unmasked,tong2022videomae,wei2022masked} extended vision transformers to the video domain, to achieve superior performance.
%
%
However, the mechanism of self-attention introduces significant computational overhead because of the similarity between all pairs of tokens needs to be calculated.

\textbf{Alternative models.} Recent work has introduced alternative models with reduced computational complexity, while maintaining the advantages of self-attention~\cite{lu2021soft,ren2021combiner,zhu2021long}.
SOFT~\cite{lu2021soft} uses Gaussian kernel functions to replace the dot-product similarity, which enables a full self-attention matrix to be approximated with a low-rank matrix decomposition.
Combiner~\cite{ren2021combiner} employs a structured factorization to approximate full self-attention, realizing low computation and memory complexity.

Peng et al.~\cite{peng2023rwkv} propose the Receptance Weighted Key Value (RWKV) architecture that combines self-attention training with an efficient recurrent neural network (RNN) inference using a linear attention mechanism. Through parallel computation, a lower, constant-level computational and memory complexity is achieved.
RetNet~\cite{sun2023retentive} includes another variant of self-attention, by dividing the input into multiple chunks. Within each chunk, the self-attention mechanism can be computed in parallel, while information is transmitted between chunks based on an RNN.

\textbf{State-space models}. The S4 model completely abandons self-attention and, instead, builds upon a state space model~\cite{gu2021efficiently}. Instead of calculatings a similarity matrix by performing matrix multiplications for pairs of tokens, it enables the network to directly learn a global high-order polynomial projection operator (HiPPO) matrix to handle relations between tokens. 
Additionally, for the simultaneous input of multiple tokens, S4 proposes a convolutional processing approach, enabling parallel training and thereby accelerating the training process.

Based on S4, Mamba~\cite{gu2023mamba} proposes a selection mechanism where, for each input token, a unique HiPPO matrix~\cite{gu2020hippo} is generated. This allows the model to selectively process input tokens, enabling it to focus on or ignore specific inputs. Due to Mamba's strong representation ability in NLP, and linear-time complexity, it has garnered attention as a promising alternative to transformers. In the computer vision domain, researchers have proposed Vision Mamba~\cite{zhu2024vision} and VMamba~\cite{liu2024vmamba} for tasks such as image classification and object detection.

In the video domain, VideoMamba~\cite{li2024videomamba} has been proposed. However, its performance is lower than expected, with limited understanding of the causes. We argue that a systematic, mathematical analysis of Mamba from the perspective of self-attention could reveal shortcomings of Mamba's inner workings. Better understanding of these limitations allow us to develop improvements, and to close the accuracy performance gap with transformers, while enjoying the efficiency of Mamba.

\section{Theoretical Analysis}
\label{sec:theory}

First, we revisit Mamba from the perspective of self-attention. Then, we analyze its limitations for video understanding. We propose VideoMambaPro to address these limitations in Section~\ref{sec:videomambapro}.

\subsection{Mamba from the perspective of self-attention}
\label{subsec:theory_self-attention}

\textbf{Self-attention}. Given an input sequence $ \bm{\mathit{X}} := \left [\bm{\mathit{x_1}}, \cdots, \bm{\mathit{x_N}} \right ] \in \mathbb{R}^{N \times D_x}$ of $N$ feature vectors of depth $D_x$, self-attention~\cite{vaswani2017attention,zhu2023biformer} computes the output sequence $\mathbf{Y}$ from $\bm{\mathit{X}}$ following two steps:

\textbf{Step 1: Similarity matrix computation.} The input sequence $\bm{\mathit{X}}$ is linearly projected onto the three different subspaces query $\mathbf{Q}\in \mathbb{R}^{N \times D}$, key $\mathbf{K} \in \mathbb{R}^{N \times D}$, and value $\mathbf{V} \in \mathbb{R}^{N \times D_V}$:

\begin{equation}
    \mathbf{Q} = \bm{\mathit{X}} \mathbf{W}_Q^{\top}, \mathbf{K} = \bm{\mathit{X}} \mathbf{W}_K^{\top}, \mathbf{V} = \bm{\mathit{X}} \mathbf{W}_V^{\top}.
\end{equation}
with $\mathbf{W}_Q,\mathbf{W}_K \in \mathbb{R}^{D \times D_x}$, and $\mathbf{W}_V \in \mathbb{R}^{D_v \times D_x}$ the corresponding weight matrices. Specifically, $\mathbf{Q} := \left [\bm{\mathit{q_1}}, \cdots, \bm{\mathit{q_N}}  \right ]^{\top}$, $\mathbf{K} :=  \left [\bm{\mathit{k_1}}, \cdots, \bm{\mathit{k_N}}  \right ]^{\top}$, and $\mathbf{V} :=  \left [\bm{\mathit{v_1}}, \cdots, \bm{\mathit{v_N}}  \right ]^{\top}$ with vectors $\bm{\mathit{q_i}}, \bm{\mathit{k_i}}, \bm{\mathit{v_i}}$ for $i = 1, \cdots, N$ the query, key, and value vectors, respectively, for input $i$. Based on $\mathbf{Q}$ and $\mathbf{K}$, similarity matrix $\mathbf{S} \in \mathbb{R}^{N \times N}$ contains the correlations between all query and key vectors, with a softmax function applied to each row of $\mathbf{S}$:
\begin{equation}
    \mathbf{S} = \text{softmax}(\mathbf{Q}\mathbf{K}^{\top}/\sqrt{D}).
\end{equation}
Each component $s_{ij}$ ($i, j = 1,\cdots,N$) represents the similarity score between $\bm{\mathit{q_i}}$ and $\bm{\mathit{k_j}}$.

\textbf{Step 2: Output computation.} Output sequence $\mathbf{Y} := \left [\bm{\mathit{y_1}}, \cdots, \bm{\mathit{y_N}} \right ]^{\top} $ is then calculated based on $\mathbf{S}$ as:
\begin{equation}
    \mathbf{Y} = \mathbf{S} \mathbf{V}.
\label{equa:self-attention result Y}
\end{equation}

It follows that each output vector $\bm{\mathit{y_i}}$ ($i = 1,\cdots,N$) can be written in vector form as:
\begin{equation}
    \bm{\mathit{y_i}} = \sum_{j=1}^{N}s_{ij}\bm{\mathit{v_j}}.
\label{equa:self-attention result yi}
\end{equation}

Any output vector $\bm{\mathit{y_i}}$ is a linear combination of vectors $\bm{\mathit{v_j}} (j = 1,\cdots,N)$, with similarity score $s_{ij}$ serving as coefficient. The larger the similarity score, the greater the influence of $\bm{\mathit{v_j}}$ on output $\bm{\mathit{y_i}}$~\cite{shen2022effects}.

\textbf{Mamba.} State Space Models (SSMs) serve as the foundation of Mamba~\cite{gu2023mamba}. They are based on continuous systems that map 1D functions or sequences, $x(t) \in \mathbb{R}^{L} \rightarrow y(t) \in \mathbb{R}^{L}$ to output sequences $y(t)$ through a hidden state $h(t) \in \mathbb{R}^{N}$. Formally, SSM implements the mapping as\footnote{The original SSM~\cite{gu2021efficiently} employs $h'(t) = Ah(t) + Bx(t)$, with $h(t)$ the hidden state from previous time step $t-1$, and $h'(t)$ the updated current hidden state, replacing $h(t)$. Considering this approach may lead to ambiguity, we have adopted the updated description.}:
\begin{equation}
    h(t) = \mathbf{A}h(t-1) + \mathbf{B}x(t),
\end{equation}
\begin{equation}
    y(t) = \mathbf{C}h(t)
\end{equation}
where $\mathbf{A} \in \mathbb{R}^{N \times N}$ is the evolution matrix of the system, and $\mathbf{B} \in \mathbb{R}^{N \times 1}$, $ \mathbf{C}\in \mathbb{R}^{N \times 1}$ are the projection matrices. Often, the input is discrete rather than a continuous function $x(t)$. Therefore, Mamba performs discretization, effectively creating a discrete version of the continuous system. A timescale parameter $\mathbf{\Delta}$ is used to transform the continuous parameters $\mathbf{A} and \mathbf{B}$ into their discrete counterparts $\mathbf{\overline{A}}, \mathbf{\overline{B}}$, and the transformation typically employs the zero-order hold method~\cite{zhang2007comparison}. 
This process is expressed as:
\begin{equation}
    \mathbf{\overline{A}} = \text{exp}(\mathbf{\Delta} \mathbf{A}),
\label{equa:discrete A}
\end{equation}
\begin{equation}
    \mathbf{\overline{B}} = (\mathbf{\Delta} \mathbf{A})^{-1} (\text{exp}(\mathbf{\Delta} \mathbf{A}) - \mathbf{I}) \cdot \mathbf{\Delta} \mathbf{B},
\label{equa:discrete B}
\end{equation}
\begin{equation}
    h_t = \mathbf{\overline{A}}h_{t-1} + \mathbf{\overline{B}}x_t,
\label{equa:discrete ht}
\end{equation}
\begin{equation}
    y_t = \mathbf{C}h_t.
\label{equa:discrete yt}
\end{equation}

Considering that parameters $\mathbf{\overline{A}},\mathbf{\overline{B}},\mathbf{C}$ in the original SSM are independent of the input data $x(t)$ and cannot be tailored to specific input data, Mamba employs a Selective Scan Mechanism as its core operator. More precisely, three functions $S_B(x), S_C(x), S_\Delta(x)$ are introduced to associate parameters $\mathbf{\overline{B}}, \mathbf{C}, and \mathbf{\Delta}$ in Equations~\ref{equa:discrete A}--\ref{equa:discrete yt} to the input data $x$. 
Based on $S_\Delta(x)$, $\mathbf{\overline{A}}$ can also be associated with the input data $x$. For example, given the input $x_1$, functions $S_\Delta(x)$ will produce the corresponding $\mathbf{\overline{A}}_1$ based on Equation \ref{equa:discrete A}, and functions $S_B(x)$ and $S_\Delta(x)$ will produce the corresponding $\mathbf{\overline{B}}_1$ based on Equation \ref{equa:discrete B}. $\mathbf{\overline{C}}_1$ is obtained based on function $S_C(x)$.
Following Equations~\ref{equa:discrete ht} and ~\ref{equa:discrete yt}, we analyze the process to obtain output sequence $\mathbf{Y}$ when given an input sequence $ \bm{\mathit{X}} := \left [\bm{\mathit{x_1}}, \cdots, \bm{\mathit{x_N}}  \right ] \in \mathbb{R}^{N \times D_x}$ of $N$ feature vectors. Each vector's hidden state is denoted as:

\begin{flalign}
 &h_1 = \mathbf{\overline{B}}_1 \bm{\mathit{x_1}} \label{equa:h1},\\
 &h_2 = \mathbf{\overline{A}}_2 h_1+ \mathbf{\overline{B}}_2 \bm{\mathit{x_2}} \notag \\
 &\quad = \mathbf{\overline{A}}_2\mathbf{\overline{B}}_1 \bm{\mathit{x_1}}+\mathbf{\overline{B}}_2 \bm{\mathit{x_2}}, \\
 &h_3 = \mathbf{\overline{A}}_3h_2+ \mathbf{\overline{B}}_3 \bm{\mathit{x_3}} \notag \\
 &\quad = \mathbf{\overline{A}}_3\mathbf{\overline{A}}_2\mathbf{\overline{B}}_1 \bm{\mathit{x_1}}+\mathbf{\overline{A}}_3\mathbf{\overline{B}}_2 \bm{\mathit{x_2}} + \mathbf{\overline{B}}_3 \bm{\mathit{x_3}},\\
 &\cdots \notag\\
 &h_N = \mathbf{\overline{A}}_Nh_{N-1}+ \mathbf{\overline{B}}_N \bm{\mathit{x_N}} \notag\\
 &\quad = \mathbf{\overline{A}}_N\mathbf{\overline{A}}_{N-1}\cdots\mathbf{\overline{A}}_2\mathbf{\overline{B}}_1 \bm{\mathit{x_1}} + \mathbf{\overline{A}}_N\mathbf{\overline{A}}_{N-1}\cdots\mathbf{\overline{A}}_{3}\mathbf{\overline{B}}_2 \bm{\mathit{x_2}} \notag \\
 &\quad + \mathbf{\overline{A}}_N\mathbf{\overline{B}}_{N-1}\bm{\mathit{x_{N-1}}} +  \mathbf{\overline{B}}_{N}\bm{\mathit{x_{N}}}. \label{equa:hN}
\end{flalign}

Equations~\ref{equa:h1}--\ref{equa:hN} can be written in matrix form:
\begin{equation}\label{equa:matrix H}
\begin{aligned}
\mathbf{H} &=  \left [h_1, h_2, h_3,\cdots, h_N \right ]^{\top}\\
&= 
\scalebox{0.72}{$
\begin{bmatrix}
\mathbf{\overline{B}}_1                           & 0                         & 0 & \cdots &   0\\
\mathbf{\overline{A}}_2 \mathbf{\overline{B}}_1   & \mathbf{\overline{B}}_2   & 0 & \cdots &   0\\
\mathbf{\overline{A}}_3\mathbf{\overline{A}}_2\mathbf{\overline{B}}_1 & \mathbf{\overline{A}}_3\mathbf{\overline{B}}_2 & \mathbf{\overline{B}}_3 & \cdots &   0 \\
\vdots & \vdots & \vdots & \ddots &\vdots\\
\mathbf{(\prod \limits_{j=N}^2 \overline{A}_j)} \mathbf{\overline{B}}_1 &  \mathbf{(\prod \limits_{j=N}^3 \overline{A}_j)} \mathbf{\overline{B}}_2&  
\mathbf{(\prod \limits_{j=N}^4 \overline{A}_j)} \mathbf{\overline{B}}_3&
\cdots & \mathbf{\overline{B}}_N
\end{bmatrix}
\begin{bmatrix}
\bm{\mathit{x_1}}\\
\bm{\mathit{x_2}} \\
\bm{\mathit{x_3}}\\
\vdots \\
\bm{\mathit{x_N}}
\end{bmatrix}
$}.
\end{aligned}
\end{equation}

For output sequence $\mathbf{Y} := \left [\bm{\mathit{y_1}}, \cdots, \bm{\mathit{y_N}} \right ]^{\top} $, each vector $\bm{\mathit{y_i}}$ ($i = 1, \cdots, N$) can be expressed as:
\begin{flalign}
    &\bm{\mathit{y_N}} = \mathbf{\overline{C}}_Nh_N \label{equa:yN},
\end{flalign}
and in matrix form as:
\begin{equation}\label{equa:matrix Y}
\mathbf{Y} = \begin{bmatrix}
\mathbf{\overline{C}}_1 & 0 & 0 &\cdots & 0\\
0 & \mathbf{\overline{C}}_2 & 0 &\cdots & 0\\
0 & 0 & \mathbf{\overline{C}}_3 & \cdots & 0\\
\vdots & \vdots &\vdots& \ddots & \vdots \\
0 & 0 & 0 &\cdots &   \mathbf{\overline{C}}_N\\
\end{bmatrix}
\begin{bmatrix}
h_1\\
h_2 \\
h_3 \\
\vdots \\
h_N 
\end{bmatrix}
= \mathbf{\overline{C}} h.
\end{equation}

By substituting Equation~\ref{equa:matrix H} into Equation~\ref{equa:matrix Y}, we obtain:
\begin{equation}\label{equa:matrix Y in full expression}
\scalebox{1.0}{$
\begin{aligned}
\mathbf{Y} =\mathbf{\overline{C}}
\scalebox{0.68}{$
\begin{bmatrix}
\mathbf{\overline{B}}_1                           & 0                         & 0 & \cdots &   0\\
\mathbf{\overline{A}}_2 \mathbf{\overline{B}}_1   & \mathbf{\overline{B}}_2   & 0 & \cdots &   0\\
\mathbf{\overline{A}}_3\mathbf{\overline{A}}_2\mathbf{\overline{B}}_1 & \mathbf{\overline{A}}_3\mathbf{\overline{B}}_2 & \mathbf{\overline{B}}_3 & \cdots &   0 \\
\vdots & \vdots & \vdots & \ddots &\vdots\\
\mathbf{(\prod \limits_{j=N}^2 \overline{A}_j)} \mathbf{\overline{B}}_1 &  \mathbf{(\prod \limits_{j=N}^3 \overline{A}_j)} \mathbf{\overline{B}}_2&  
\mathbf{(\prod \limits_{j=N}^4 \overline{A}_j)} \mathbf{\overline{B}}_3&
\cdots & \mathbf{\overline{B}}_N
\end{bmatrix}
\begin{bmatrix}
\bm{\mathit{x_1}}\\
\bm{\mathit{x_2}} \\
\bm{\mathit{x_3}}\\
\vdots \\
\bm{\mathit{x_N}}
\end{bmatrix}
$}
\end{aligned},
$}
\end{equation}
which can be expressed as:
\begin{equation}
    \mathbf{Y} = \mathbf{\overline{C}} (\mathbf{M} \bm{\mathit{X}}),
\label{equa:Matrix Y in CAX}
\end{equation}
where $\mathbf{M}$ represents the second term on the right-hand side of Equation~\ref{equa:matrix Y in full expression}. Recall from Equation~\ref{equa:self-attention result Y} that the result $\mathbf{Y}$ obtained by self-attention processing can be expressed as:
\begin{equation}
    \mathbf{Y} = \mathbf{S}\mathbf{V} =  (\mathbf{S} \bm{\mathit{X}}) \mathbf{W}_V^{\top}
\label{equa: mamba result Y}
\end{equation}

From the perspective of self-attention, by comparing Equations~\ref{equa:Matrix Y in CAX} and \ref{equa: mamba result Y}, the essence of Mamba is to generate a matrix $\mathbf{M}$ similar to similarity matrix $\mathbf{S}$, such that the result of $\mathbf{M}\bm{\mathit{X}}$ is based on the correlation between vectors of $\bm{\mathit{X}}$. Although the final result of $\mathbf{M}\bm{\mathit{X}}$ is left multiplied by a mapping matrix $\mathbf{C}$, while the result of $\mathbf{S}\bm{\mathit{X}}$ is right multiplied by a mapping matrix $\mathbf{W}_V^{\top}$, the geometric meaning of the two are the same.

\subsection{Limitations of Mamba in video understanding}
\label{subsec:theory_limitations}

From the perspective of self-attention, the concept of Mamba is similar: both use similarity matrices. We now analyze the differences between the similarity matrices of Mamba and self-attention, and discuss the limitations of Mamba in the context of the video understanding task.


\textbf{Limitation 1: Historical decay.} Matrix $\mathbf{M}$ in Equation~\ref{equa:Matrix Y in CAX} corresponds to the second right-hand term in Equation~\ref{equa:matrix Y in full expression}, which is a lower triangular matrix of the form:
\begin{equation}\label{equa:matrix A}
\scalebox{1.0}{$
\mathbf{M} = \begin{bmatrix}
m_{11}                           & 0                         & 0 & \cdots &   0\\
m_{21}   & m_{22}   & 0  & \cdots &   0\\
m_{31}   & m_{32}   & m_{33}  & \cdots &   0 \\
\vdots & \vdots & \vdots & \ddots &\vdots \\
m_{N1} & m_{N2} &  m_{N3}& \cdots & m_{NN}
\end{bmatrix}.
$}
\end{equation}

By comparing $\mathbf{M}$ with matrix $\mathbf{S}$ in self-attention, we find that outputs in Mamba favor more recent information, because the more weights are zero, the earlier the token is observed. For example, for input $\left [\bm{\mathit{x_1}}, \bm{\mathit{x_2}},\bm{\mathit{x_3}} \right]$, the output $\mathbf{M}\bm{\mathit{x_1}}$ in Mamba is $m_{11} \bm{\mathit{x_1}}$ while the output $\mathbf{S}\bm{\mathit{x_1}}$ is $s_{11}\bm{\mathit{x_1}} + s_{12}\bm{\mathit{x_2}} + s_{13} \bm{\mathit{x_3}}$ in self-attention. This indicates that, in Mamba, the influence of earlier observed tokens on the final result is greatly diminished. We refer to this limitation as historical decay.

In the NLP domain, more recent dialogue information often has more impact on the final judgment, so this effect is acceptable. However, in the computer vision domain, the order of the tokens has less meaning. Previous works such as Vision Mamba~\cite{zhu2024vision} and VMamba~\cite{liu2024vmamba} have partly mitigated this issue by processing the token sequence in both forward and backward directions. This produces better results but no work has explained why this is effective.

When processing bi-directionally, the results generated from input forward tokens $\left [\bm{\mathit{x_1}}, \cdots,\bm{\mathit{x_N}} \right ]$, denoted as $\mathbf{M}_{f}\bm{\mathit{X}}$, and the results generated from input backward tokens $\left [\bm{\mathit{x_N}}, \cdots,\bm{\mathit{x_1}} \right ]$, denoted as $\mathbf{M}_{b}\bm{\mathit{X}}$, are linearly combined to generate the final result $\mathbf{M}_{bi}\bm{\mathit{X}}$ with $\mathbf{M}_{bi}$ a dense matrix. As a result, the influence of historical information on the result is increased, consequently leading to better results.

For example, for the input tokens $\left [\bm{\mathit{x_1}}, \bm{\mathit{x_2}},\bm{\mathit{x_3}} \right ]$, $\mathbf{M}_{f}\bm{\mathit{X}}$ and $\mathbf{M}_{b}\bm{\mathit{X}}$ can be expressed as:
\begin{equation}\label{equa:Af}
\mathbf{M}_{f}\bm{\mathit{X}} =
\begin{bmatrix}
f_{11} & 0 & 0 \\
f_{21} & f_{22} & 0 \\
f_{31} & f_{32} & f_{33}
\end{bmatrix}
\begin{bmatrix}
\bm{\mathit{x_1}}\\
\bm{\mathit{x_2}}\\
\bm{\mathit{x_3}}
\end{bmatrix}
=
\begin{bmatrix}
h_{1f}\\
h_{2f}\\
h_{3f}
\end{bmatrix},
\end{equation}

\begin{equation}\label{equa: Ab}
\mathbf{M}_{b}\bm{\mathit{X}} =
\begin{bmatrix}
b_{33} & 0 & 0 \\
b_{23} & b_{22} & 0 \\
b_{13} & b_{12} & b_{11}
\end{bmatrix}
\begin{bmatrix}
\bm{\mathit{x_3}}\\
\bm{\mathit{x_2}}\\
\bm{\mathit{x_1}}
\end{bmatrix}
=
\begin{bmatrix}
h_{3b}\\
h_{2b}\\
h_{1b}
\end{bmatrix},
\end{equation}
where $f_{ij}$ represents the similarity score during the forward process, and $b_{ij}$ is the similarity score in the backward direction. After bi-directional computation, with the outputs linearly combined, the results are expressed as:
\begin{equation}\label{equa:h1 to h3}
\scalebox{0.98}{$
\begin{aligned}
h_1 &= h_{1f}+h_{1b} = f_{11}\bm{\mathit{x_1}} + b_{13}\bm{\mathit{x_3}}+b_{12}\bm{\mathit{x_2}}+b_{11}\bm{\mathit{x_1}} \\
h_2 &= h_{2f}+h_{2b} = f_{21}\bm{\mathit{x_1}} + f_{22}\bm{\mathit{x_2}}+ b_{23}\bm{\mathit{x_3}}+b_{22}\bm{\mathit{x_2}}\\
h_3 &= h_{3f}+h_{3b} = f_{31}\bm{\mathit{x_1}} + f_{32}\bm{\mathit{x_2}} + f_{33}\bm{\mathit{x_3}} + b_{33}\bm{\mathit{x_3}} 
\end{aligned}
$}
\end{equation}
   
We can write Equation~\ref{equa:h1 to h3} in matrix form:
\begin{equation}\label{equa:h1 to h3 in Abi format}
\scalebox{0.98}{$
\begin{aligned}
\begin{bmatrix}
h_1\\
h_2\\
h_3\\
\end{bmatrix}
&=
\begin{bmatrix}
f_{11}+b_{11} & b_{12} & b_{13}\\
f_{21} & f_{22}+b_{22} & b_{23}\\
f_{31} & f_{32} & f_{33}+b_{33}\\
\end{bmatrix}
\begin{bmatrix}
\bm{\mathit{x_1}}\\
\bm{\mathit{x_2}}\\
\bm{\mathit{x_3}}\\
\end{bmatrix}\\
&= \mathbf{M}_{bi}
\begin{bmatrix}
\bm{\mathit{x_1}}\\
\bm{\mathit{x_2}}\\
\bm{\mathit{x_3}}\\
\end{bmatrix}.
\end{aligned}
$}
\end{equation}

The bi-directional computation transforms the original matrix $\mathbf{M}$ from a lower triangular matrix to a dense matrix $\mathbf{M}_{bi}$, thereby capturing more historical information and effectively avoiding the historical decay. When extending to the case of $N$ input tokens $\left [\bm{\mathit{x_1}}, \cdots,\bm{\mathit{x_N}} \right ]$, $\mathbf{M}_{bi}$ can be written as
\begin{equation}
\scalebox{0.78}{$
\mathbf{M}_{bi} =
\begin{bmatrix}
f_{11}+b_{11} & b_{12} & b_{13} & \cdots & b_{1N}\\
f_{21} & f_{22}+b_{22} & b_{23} & \cdots & b_{2N}\\
f_{31} & f_{32} & f_{33}+b_{33} & \cdots & b_{3N}\\
\vdots & \vdots & \vdots        & \ddots & \vdots \\
f_{N1} & f_{N2} & f_{N3}        & \cdots & f_{NN}+b_{NN}\\
\end{bmatrix}
$}.
\end{equation}

The diagonal elements of $\mathbf{M}_{bi}$ contain duplicates of the similarity between a token and itself. For example, $f_{33}$ and $b_{33}$ each represent the similarity between token $\bm{\mathit{x_3}}$ and itself. Consequently, the similarity is effectively doubled which weakens the association with other tokens. One possible approach is to adjust $\mathbf{M}_{f}$ and $\mathbf{M}_{b}$ using a weight coefficient $z$ through a linear combination. However, learning such a parameter $z$ that weakens the diagonal elements without affecting other elements might be challenging.

\begin{figure*}[t]
  \centering
  \includegraphics[width=1.0\linewidth]{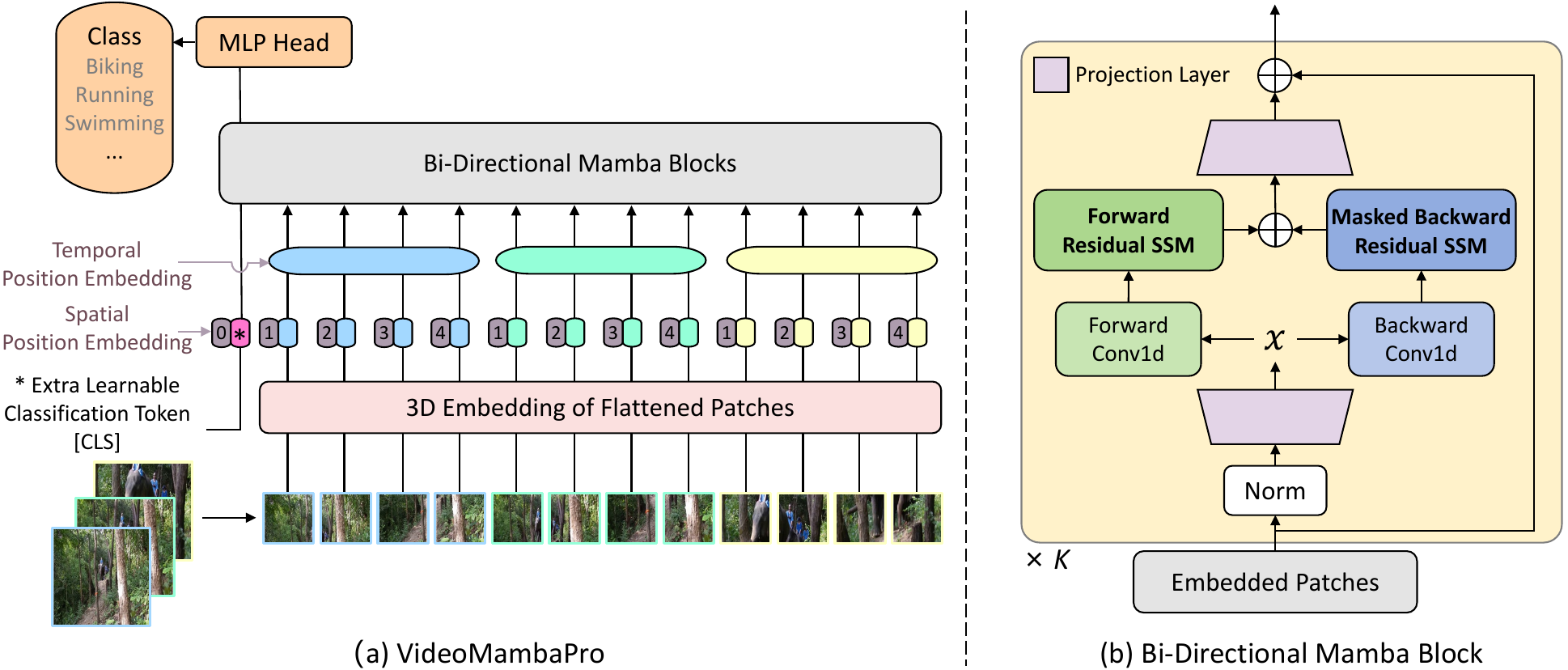}
   \caption{(a) Framework of VideoMambaPro with $K$ bi-directional Mamba blocks. (b) In each bi-directional Mamba block, we employ forward residual SSM and masked backward Residual SSM.}
   \label{fig:1}
\end{figure*}

\textbf{Limitation 2: Element contradiction.} 
By analyzing the non-zero elements $m_{ij}$ in $\mathbf{M}$ of Equation~\ref{equa:matrix Y in full expression}, it can be summarized that:
\begin{equation}\label{limit2}
    m_{ij}=\mathbf{\overline{A}}_im_{i-1, j}
\end{equation}
After multiple iterations, the above equation results in implicit consideration of the correlation between previous tokens and token $j$ when computing the correlation between token $i$ and token $j$. As a result, $m_{ij}$ exhibits stronger contextual dependencies compared to the elements $s_{ij}$ in the matrix $\mathbf{S}$. This might explain why Mamba achieves better performance than transformers in the field of NLP. While this is advantageous in the NLP domain, for the computer vision domain, input tokens often lack semantic connections. The consideration of the influence of other tokens on each element can lead to significant drawbacks.

We often observe an interleaved token structure when processing images. Tokens that ``belong together'' might not be subsequently processed. For example, in an image classification task, input tokens $\left [\bm{\mathit{x_1}}, \bm{\mathit{x_2}},\bm{\mathit{x_3}} \right ]$ might represent image regions $\left[\textit{dog}, \textit{other}, \textit{dog}\right]$. Ideally, $m_{31}$ should be a high value and $m_{21}$ should be low. According to Equation~\ref{limit2}, $m_{31}=\mathbf{\overline{A}}_3m_{21}$, which requires the network to set $\mathbf{\overline{A}}_3$ to a high value to meet the requirement on $m_{31}$. However, in doing so, $m_{32}=\mathbf{\overline{A}}_3m_{22}$ would also become larger because $m_{22}$ is also high. But, theoretically, $m_{32}$ should be low. This leads to an element contradiction. Especially for video understanding, such contradictions are common because most video regions contain background and other irrelevant information, making relevant tokens sparse. Consequently, the performance of Mamba applied to video analysis tasks is underwhelming~\cite{li2024videomamba,li2024spikemba,zhang2024survey}.

\section{VideoMambaPro}
\label{sec:videomambapro}

We propose two adaptations to VideoMamba~\cite{li2024videomamba} to address the two identified limitations: historical decay and element contradiction. The resulting architecture is termed VideoMambaPro (VMP). With minor adjustments, our adaptations can also be applied to related Mamba models.

To address historical decay, we keep the result of $\mathbf{M}_{f}\bm{\mathit{X}}$ unchanged but we use masked  computation during the backward process. Specifically, we assign a mask to the diagonal elements of $\mathbf{M}_{b}$, setting their values to $0$, and then proceed with the calculations in Equations~\ref{equa:matrix A}--\ref{equa:h1 to h3 in Abi format}. We thus eliminate the duplicate similarity on the diagonal, without affecting other elements. The final $\mathbf{M}_{bi}$ is expressed as: 
\begin{equation}
M_{bi}=
\begin{bmatrix}
f_{11} & b_{12} & b_{13} & \cdots & b_{1N}\\
f_{21} & f_{22} & b_{23} & \cdots & b_{2N}\\
f_{31} & f_{32} & f_{33} & \cdots & b_{3N}\\
\vdots & \vdots & \vdots        & \ddots & \vdots \\
f_{N1} & f_{N2} & f_{N3}        & \cdots & f_{NN}\\
\end{bmatrix}.
\end{equation}

To solve the element contradiction issue, we propose residual SSM, inspired by residual connections, to distribute the requirement for $\mathbf{\overline{A}}_i$ in $m_{ij}$ across multiple $\mathbf{\overline{A}}_i$. This helps to avoid contradictions caused by interleaved sequence structures. For example, for our previous example input sequence $\left [\bm{\mathit{x_1}}, \bm{\mathit{x_2}},\bm{\mathit{x_3}} \right ]$, which represents regions $\left[\textit{dog}, \textit{other}, \textit{dog}\right]$, we let $m_{31}=\mathbf{\overline{A}}_3m_{21}+\mathbf{\overline{A}}_3$. This way, the requirement for a single $\mathbf{\overline{A}}_3$ can be split into two parts, thus avoiding contradictions. This can be expressed as:
\begin{equation}
    m_{ij}=\mathbf{\overline{A}}_im_{i-1, j}+\mathbf{\overline{A}}_i
\end{equation}

With these two solutions, we propose VideoMambaPro, based on VideoMamba~\cite{li2024videomamba} and illustrated in Figure~\ref{fig:1}. Given input video $\bm{\mathit{X}}^v \in \mathbb{R}^{3 \times T \times H \times W}$, we first use a 3D convolution with a $1\times16\times16$ kernel to convert $\bm{\mathit{X}} ^v$ into $L$ non-overlapping patch-wise tokens $\bm{\mathit{X}} ^p \in \mathbb{R}^{L \times C}$, where $L = t \times h \times w$ $(t = T, h = \frac{H}{16}, w = \frac{W}{16}$. 
Because SSM is sensitive to token positions, and in line with VideoMamba, we include a learnable spatial position embedding $\bm{\mathit{p_s}} \in \mathbb{R}^{(hw+1) \times C}$ and a temporal position embedding $\bm{\mathit{p_t}} \in \mathbb{R}^{t \times C}$. Input tokens $\bm{\mathit{X}}$ are expressed as:
\begin{equation}
   \bm{\mathit{X}} = [\bm{\mathit{X}}_{cls}, \bm{\mathit{X}}] + \bm{\mathit{p_s}} + \bm{\mathit{p_t}},
\end{equation}
where $\bm{\mathit{X}}_{cls}$ is a learnable classification token positioned at the start of the sequence. Input tokens $\bm{\mathit{X}}$ pass through $K$ Mamba blocks, and the final layer's [CLS] token is used for classification, after normalization and linear projection.

\section{Experiments} \label{sec:experiments}

\subsection{Experimental setup}
\textbf{Datasets.} We evaluate VideoMambaPro on five video benchmarks:
(a) Kinetics-400 (K400,~\cite{carreira2017quo}) comprises $\sim$240K training and $\sim$20K validation videos, each with an average duration of 10 seconds and categorized into 400 classes.
(b) Something-Something V2 (SSv2,~\cite{goyal2017something}) includes $\sim$160K training and $\sim$20K validation videos with an average duration of 4 seconds, and 174 motion-centric classes.
(c) UCF-101~\cite{soomro2012ucf101} is a relatively small dataset, consisting of $\sim$9.5K training and $\sim$3.5K validation videos.
(d) HMDB51~\cite{kuehne2011hmdb} is also a compact video dataset, containing $\sim$3.5K training and $\sim$1.5K validation videos.
(e) AVA~\cite{gu2018ava} is a dataset for spatio-temporal localization of human actions with $\sim$211k and $\sim$57k validation video segments.

\noindent
\textbf{Implementation.} In line with VideoMamba, we introduce three models with increasing embedding dimension and number of bi-directional Mamba blocks $K$: Tiny, Small, and Middle (details in supplementary material). To compare with VideoMamba, we pre-train VideoMambaPro on ImageNet-1K (IN-1K). On K400, we also pre-train with IN-1K, fine-tune on the training set and report on the validation set. For K400, we also report on the larger $336^2$ input size.
During pre-training, we follow DeiT~\cite{touvron2021training} by applying a center crop to obtain the $224^2$ sized images. We apply random cropping, random horizontal flipping, label-smoothing regularization, mix-up, and random erasing as data augmentations. We use AdamW~\cite{loshchilov2017decoupled} with a momentum of 0.9, a batch size of 1024, and a weight decay of 0.05. We employ a cosine learning rate schedule during training, $1 \times 10^{-3}$ initial learning rate over 300 epochs. The fine-tuning settings follow VideoMAE~\cite{tong2022videomae}. We resize frames to $224^2$, and use AdamW with a momentum of 0.9 and a batch size of 512. Details in the supplementary materials.

\begin{table}[htb]
  \centering
  \resizebox{\linewidth}{!}{
  \begin{tabular}{llccrrrr}
    \toprule
    Method & Pre-train &Input & Crops & Param & FLOP & Top1 & Top5 \\
    \midrule
    MViTv1-B~\cite{fan2021multiscale}&  & $32\times224^2$ &5$\times$1 & 37M &350G &80.2 &94.4\\
    MViTv2-S~\cite{li2022mvitv2}&  & $16\times224^2$ &5$\times$1 & 35M &320G&81.0 &94.6\\ 
    Uniformer-S~\cite{li2022uniformer}& IN-1K& $16\times224^2$ &4$\times$1 & 21M &168G &80.8 &94.7 \\
    Uniformer-B~\cite{li2022uniformer}& IN-1K& $16\times224^2$ &4$\times$1 & 50M &388G &82.0 &95.1 \\
    Uniformer-B~\cite{li2022uniformer}& IN-1K& $32\times224^2$ &4$\times$3 & 50M &3.1T &83.0 &95.4 \\
    STAM~\cite{sharir2021image} &IN-21K & $64\times224^2$  &1$\times$1 &121M &1.0T &79.2 &-\\
    TimeSformer-L~\cite{bertasius2021space} &IN-21K& $96\times224^2$ &1$\times$3  & 121M &7.1T  &80.7&94.7 \\
    ViViT-L~\cite{arnab2021vivit} &IN-21K& $16 \times 224^2$ &4$\times$3 & 311M &47.9T & 81.3 & 94.7\\  
    Mformer-HR~\cite{patrick2021keeping} & IN-21K& $16 \times 336^2$ & 10$\times$3 & 311M & 28.8T& 81.1 & 95.2\\ 
        
    VideoMAE-H~\cite{tong2022videomae}& IN-21K & $16\times224^2$ &5$\times$3 & 633M &17.9T&86.6 &97.1\\ 
    X-CLIP-L/14~\cite{ni2022expanding} &CLIP-400M & $16\times336^2$  &4$\times$3 & 453M &37.0T &87.7 &---\\
    MTV-H~\cite{yan2022multiview} &60M$^1$ & $32\times224^2$  &4$\times$3 &1120M &44.5T &89.1 &98.2\\
    InternVideo-1B~\cite{wang2022internvideo} &412M$^2$ & $64 \times 224^2$ &16$\times$4 & 1300M &86.2T & 91.1 & \textbf{98.9} \\  
    InternVideo2-1B~\cite{wang2024internvideo2} & 414M$^3$ & $16 \times 224^2$ & 16$\times$4 & 1000M & --- & 91.6 & --- \\            
    InternVideo2-6B~\cite{wang2024internvideo2} & 414M$^3$ & $16 \times 224^2$ & 16$\times$4 & 5903M & --- & \textbf{92.1} & --- \\   
        
    \midrule
    VideoMamba-Ti &IN-1K & $32 \times 224^2$ &4$\times$3 & 7M & 0.4T & 78.8 & 93.9 \\
    VideoMamba-Ti  &IN-1K & $64 \times 384^2$ &4$\times$3 & 7M & 2.4T & 80.3 & 94.8\\
    VideoMamba-S  &IN-1K & $32 \times 224^2$ &4$\times$3 & 26M & 1.6T & 81.5 & 95.2\\
    VideoMamba-S  &IN-1K & $64 \times 384^2$ &4$\times$3 & 26M & 4.7T & 82.7 & 95.6\\
    VideoMamba-M &IN-1K & $32 \times 224^2$  &4$\times$3 & 74M &4.8T &82.4 &95.7 \\
    VideoMamba-M  &IN-1K & $64 \times 384^2$ &4$\times$3 & 74M & 28.4T &83.3 & 96.1\\
    VideoMambaPro-Ti &IN-1K & $32 \times 224^2$ &4$\times$3 & 7M & 0.4T & 81.6 & 95.9 \\
    VideoMambaPro-Ti  &IN-1K & $64 \times 384^2$ &4$\times$3 & 7M & 2.2T & 83.3 & 96.1 \\
    VideoMambaPro-S &IN-1K & $32 \times 224^2$ &4$\times$3 & 25M & 1.6T & 83.3 & 96.0\\
    VideoMambaPro-S  &IN-1K & $64 \times 384^2$ &4$\times$3 & 25M &4.4T & 84.5 & 96.6 \\
    VideoMambaPro-M &IN-1K & $32 \times 224^2$ &4$\times$3 & 72M & 4.7T & 84.0 & 96.4\\
    VideoMambaPro-M  &IN-1K & $64 \times 384^2$ &4$\times$3 & 72M &27.0T &\textbf{85.0} &\textbf{96.7}\\
    \bottomrule
  \end{tabular}
  }
  \caption{Performance on K400. Top part of the table are Transformer models, bottom part are Mamba models. We report crops (temporal $\times$ spatial) and FLOPs for inference. ---: not reported.
  \\$^1$ IN-21K+WTS
  \\$^2$ CLIP-400M+WebVid+HowTo+K710+SSv2+AVA2.2+more.
  \\$^3$ LAION-300M+KMash+WebVid+InternVid+LLaVA+more.}
  \label{tab:kinetics400}
\end{table}

\subsection{Comparison with state-of-the-art}
\textbf{K400.} Results appear in Table~\ref{tab:kinetics400}. Compared to VideoMamba, VideoMambaPro has slightly fewer parameters and FLOPs. This is primarily because VideoMamba employs an additional projection layer to generate the weight coefficient $z$ to adjust $\mathbf{A}_{f}$ and $\mathbf{A}_{b}$. See the supplementary materials for an architecture comparison. VideoMambaPro outperforms VideoMamba across model and input sizes. With $224^2$ inputs and pre-trained only on IN-1K, the best-performing VideoMambaPro-M achieves a top-1 accuracy of 84.0\%, 1.6\% higher than VideoMamba-M. Further comparisons appear in Section~\ref{subsec:comparison}. Increasing the input size to $336^2$ leads to a performance improvement of 1.0-1.7\%.

VideoMambaPro scores lower than the recent InternVideo2-1B~\cite{wang2024internvideo2} by 7.6\%, but was only pre-trained on IN-1K and has significantly fewer parameters (~1000M vs 72M) and inference only takes $\sim$5.5\% of the FLOPs.

\begin{table}[htb]
  \centering
  \resizebox{\linewidth}{!}{
  \begin{tabular}{llccrrrr}
    \toprule
    Method & Pre-train &Input & Crops & Param & FLOP & Top1 & Top5 \\
    \midrule
    MViTv1-B~\cite{fan2021multiscale}& K400 & $16\times224^2$ &1$\times$3 & 37M &213G &64.7 &89.2\\
    MViTv1-B~\cite{fan2021multiscale}& K400 & $32\times224^2$ &1$\times$3 & 37M &510G &67.1 &90.8\\
    MViTv2-S~\cite{li2022mvitv2}& K400 & $16\times224^2$ &1$\times$3 & 35M &195G&68.2 &91.4\\ 
    MViTv2-B~\cite{li2022mvitv2}& K400 & $32\times224^2$ &1$\times$3 & 51M &675G&70.5 &92.7\\  
    
    Uniformer-S~\cite{li2022uniformer}& IN-1K+K400& $16\times224^2$ &1$\times$3 & 21M &126G &67.7 &91.4 \\
    Uniformer-B~\cite{li2022uniformer}& IN-1K+K400& $16\times224^2$ &1$\times$3 & 50M &291G &70.4 &92.8 \\
    
    TimeSformer-L~\cite{bertasius2021space} &IN-21K& $16\times224^2$ &1$\times$3  & 121M &5.1T  &62.5&- \\
    ViViT-L~\cite{arnab2021vivit} &IN-21K+K400& $16 \times 224^2$ &4$\times$3 & 311M &47.9T & 65.4 & 89.8\\  
    Mformer-HR~\cite{patrick2021keeping} & IN-21K+K400& $16 \times 336^2$ & 1$\times$3 & 311M & 3.6T& 68.1 & 91.2\\

    MaskFeat-L~\cite{wei2022masked} & IN-21K  &$64 \times 312^2$ &4×3 &218M &8.5T &75.0 &95.0 \\
    VideoMAE-L~\cite{tong2022videomae} & IN-21K  &$32 \times 224^2$  &1×3 &305M & 4.3T  &75.4 &95.2\\
    TubeViT-L~\cite{piergiovanni2023rethinking} & IN-1K  &$32 \times 224^2$  &4×3 &311M & 9.5T &76.1 &95.2 \\
    InternVideo-1B~\cite{wang2022internvideo} & See Table~\ref{tab:kinetics400} &$64 \times 224^2$ &16×4 & 1300M &86.2T &77.2 &95.9 \\
    InternVideo2-1B~\cite{wang2024internvideo2}& See Table~\ref{tab:kinetics400}& $64 \times 224^2$ & 16×4 &~1000M & — & 77.1 & —\\
    InternVideo2-6B~\cite{wang2024internvideo2}& See Table~\ref{tab:kinetics400}& $64 \times 224^2$ & 16×4 &5903M & — & 77.4 & —\\
    
    \midrule
    VideoMamba-Ti &IN-1K & $16 \times 224^2$ &2$\times$3 & 7M & 102G & 66.0 & 89.6 \\
    VideoMamba-S  &IN-1K & $16 \times 224^2$ &2$\times$3 & 26M & 408G & 67.6 & 90.9\\
    VideoMamba-M &IN-1K & $16 \times 224^2$  &4$\times$3 & 74M &2.4T &68.3 &91.4\\
    VideoMambaPro-Ti &IN-1K & $16 \times 224^2$ &2$\times$3 & 7M & 96G & 67.9 & 91.2 \\
    VideoMambaPro-S  &IN-1K & $16 \times 224^2$ &2$\times$3 & 25M & 382G & 68.8 & 91.4\\
    VideoMambaPro-M &IN-1K & $16 \times 224^2$ &4$\times$3 & 72M & 2.2T & 69.4 & 91.6\\
    \bottomrule
  \end{tabular}
  }
  \caption{Performance on SSv2. ---: not reported. Top part of the table are Transformer models, bottom part are Mamba models.}
  \label{tab:ssv2}
\end{table}

\noindent
\textbf{SSv2.} Results appear in Table~\ref{tab:ssv2}. VideoMambaPro outperforms VideoMamba by 1.1--1.9\%. It also outperforms several popular transformer models. Although InternVideo-1B~\cite{wang2022internvideo} and InternVideo2-6B~\cite{wang2024internvideo2} outperform our VideoMambaPro-M by 7.8\% and 8.0\%, respectively, they require 18.0-82 times more parameters and at least 39 times more FLOPs. Again, we expect that the performance for VideoMambaPro will increase with more pre-training.

\begin{table}[htb]
  \centering
  \resizebox{\linewidth}{!}{
  \begin{tabular}{lccc}
    \toprule
    Method & Params & UCF-101 &HMDB51 \\
    \midrule
    VideoMoCo~\cite{pan2021videomoco} & 15M &78.7  &49.2\\
    CoCLR~\cite{han2020self} & 9M & 81.4 &52.1\\
    MemDPC~\cite{han2020memory} & 32M &86.1   &54.5\\
    Vi$^2$CLR~\cite{diba2021vi2clr} & 9M &89.1 &55.7\\
    VideoMAE~\cite{tong2022videomae} & 87M& 91.3 &62.6\\
    GDT~\cite{patrick2020multimodal} &33M &95.2 &72.8 \\
    VideoMAE V2~\cite{wang2023videomae}&1050M & \textbf{99.6} & \textbf{88.1}\\
     \midrule
     VideoMamba-M &74M &88.2 &60.8 \\
     VideoMambaPro-M&72M&91.6 & 63.2 \\
    \bottomrule
  \end{tabular}
    }
    \caption{Results on UCF-101 and HMDB51.}
    \label{tab:UCF101-HMDB51}
\end{table}

\begin{table}[htb]
    \resizebox{\linewidth}{!}{
    \begin{tabular}{lcccc}
    \toprule
    Method & FLOPs &Param& mAP  \\
    \midrule
    SlowFast R101~\cite{feichtenhofer2019slowfast}   &138G& 53M &23.8\\
    VideoMAE-B~\cite{tong2022videomae}  &180G& 87M &26.7\\
    MViTv2-B~\cite{li2022mvitv2} & 225G &51M &30.5\\
    ObjectTransformer~\cite{wu2021towards}& 243G &86M &31.0\\
    MViTv2-L~\cite{li2022mvitv2} & 2828G &213M &34.4\\
    ST-MAE-H~\cite{feichtenhofer2022masked}   &1193G& 632M &36.2\\
    VideoMAE V2~\cite{wang2023videomae}   &4220G& 1050M &42.6\\
    
    

     \midrule
    VideoMamba-M~\cite{li2022uniformerv2} & 202G &74M&30.1 \\
    VideoMambaPro-M &183G  &72M &31.9\\
    \bottomrule
  \end{tabular}
  }
    \caption{Results on AVA V2.2.}
    \label{tab:AVA}
\end{table}

\noindent
\textbf{UCF-101/HMDB51/AVA V2.2.} From Table~\ref{tab:UCF101-HMDB51}, it shows that VideoMambaPro-M is competitive, and outperforms VideoMamba by 3.4\% and 1.8\% on UCF-101 and HMDB51, respectively. VideoMambaPro-M achieves 31.9 mAP on AVA V2.2, which is 10.7\% lower than VideoMAE V2~\cite{wang2023videomae} but with an order of magnitude fewer parameters and FLOPs and pre-trained only on IN-1K (see Table~\ref{tab:AVA}).

\begin{table}[htb]
\centering
\resizebox{\linewidth}{!}{
\begin{tabular}{llcc}
\toprule
Models                                    & Input     & Top-1       & Top-5       \\ \midrule
VideoMamba-M (baseline)                              & $32 \times 224^2$ & 82.4       & 95.7        \\ 
VideoMambaPro-M (w/o residual)  & $32 \times 224^2$ & 83.6 (+1.2)        & 96.0 (+0.3)        \\ 
VideoMambaPro-M (w/o masking)  & $32 \times 224^2$ & 83.0 (+1.0)        & 95.8 (+0.1)        \\ 
VideoMambaPro-M & $32 \times 224^2$ & 84.0 (+1.6) & 96.4 (+0.7) \\ \bottomrule
\end{tabular}
}
\caption{Ablation study on K400, without and without masked backward computation and elemental residual connections.}
\label{tab:ablation study}
\end{table}

\subsection{Ablation study}
We have identified two limitations that exist in VideoMamba: historical decay and element contradiction. We introduced masked backward computation and elemental residual connections to address these respective issue. Here, we analyze the impact of each solution. We use the same settings as before, with VideoMambaPro-M and pre-training on IN-1K. We summarize the performance of VideoMambaPro-M on K400 in Table~\ref{tab:ablation study}. Both solutions contribute to an improved score, and their effect is partly complementary. This indicates that the two limitations exist simultaneously in VideoMamba.

\subsection{Comparison with VideoMamba on K400}
\label{subsec:comparison}

We more thoroughly compare the differences between VideoMamba and VideoMambaPro by investigating the relative performance per class. We then present a statistical comparison between the results of both backbones.

\begin{figure}[htbp]
  \centering
  \includegraphics[width=1.0\linewidth]{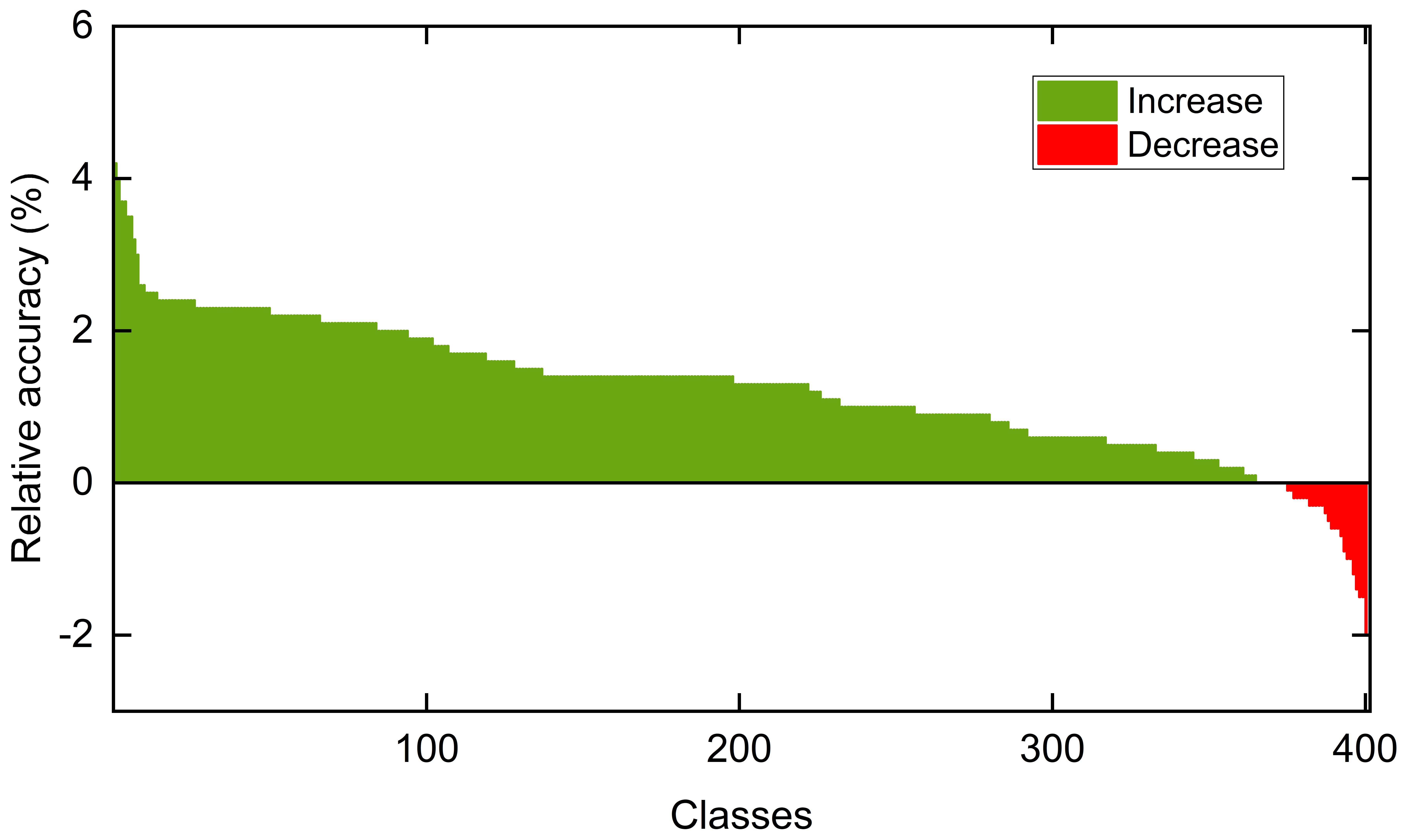}
   \caption{Relative accuracy per class on Kinetics-400 by comparing VideoMambaPro-M to a baseline VideoMamba-M. Classes sorted by relative performance.}
   \label{fig:accuracy improvement on all classes}
\end{figure}

\noindent
\textbf{Class analysis}. We compare VideoMambaPro-M with $224^2$ image size pre-trained on IN-1K to a VideoMamba-M baseline with the same settings. We show the relative performance for all classes of K400 in Figure~\ref{fig:accuracy improvement on all classes}. For over 95\% of the classes, VideoMambaPro shows improvement. Although there is a lower performance for certain classes, the decrease is typically limited.

The majority of the classes sees a $\sim$1.8\% improvement, which is substantial. For a small number of classes, VideoMambaPro performs $>$2\% better than VideoMamba. Only a fraction of the classes is negatively affected by the solutions introduced in VideoMambaPro.

\textbf{Statistical comparison}. In order to understand whether the improvements of VideoMambaPro over VideoMamba are statistically significant, we compare the results of the respective Middle models, both pre-trained on IN-1K and with a spatial input size of $224 \times 224$ and applied to K400. Other settings are also the same. For each test sample, we check whether if it correctly classified by either model. The summary of these results appears in Table~\ref{tab:stat_overview}.

\begin{table}[htbp]
\centering
\scalebox{0.9}{
\begin{tabular}{ccccc}
\toprule
                            &       & \multicolumn{3}{c}{VideoMambaPro-M} \\
\midrule
                            &       & True      & False     & Total     \\
                            \cmidrule{2-5}
\multirow{3}{*}{VideoMamba-M} & True  & 14,302     & 469       & 14,771     \\
                            & False & 1,833      &3,302    & 5,135      \\
                            \cmidrule{2-5}
                            & Total & 16,135     & 3,771      & 19,906     \\
\bottomrule                            
\end{tabular}
}
\caption{Contingency table for K400 test items for VideoMamba-M and VideoMambaPro-M. }
\label{tab:stat_overview}
\end{table}

We used the McNemar test, a non-parametric test with a single degree of freedom. Essentially, it checks whether the number of items that are incorrectly classified by VideoMambaPro-M but not VideoMamba is substantially lower than the number of items misclassified by VideoMamba but not VideoMambaPro. The test is calculated as $\chi^2 = \frac{(n_{01} - n_{10})^2}{(n_{01} + n_{10})}$ with $n_{01}$ corresponding to the number of items that were misclassified by VideoMamba but not VideoMambaPro, and $n_{10}$ the number of items that were correctly classified by VideoMamba but misclassified by VideoMambaPro. These numbers correspond to 1,833 and 469, respectively. Based on the Chi-square distribution, the resulting value of 808.2 corresponds to a significance level of $p < 0.001$. We can thus conclude that VideoMambaPro-M is statistically significantly better than VideoMamba.

Because we relied on the aggregated performance reported in papers for other methods, we cannot report statistical comparisons here.

\begin{figure}[htb]
  \centering
  \includegraphics[width=1.0\linewidth]{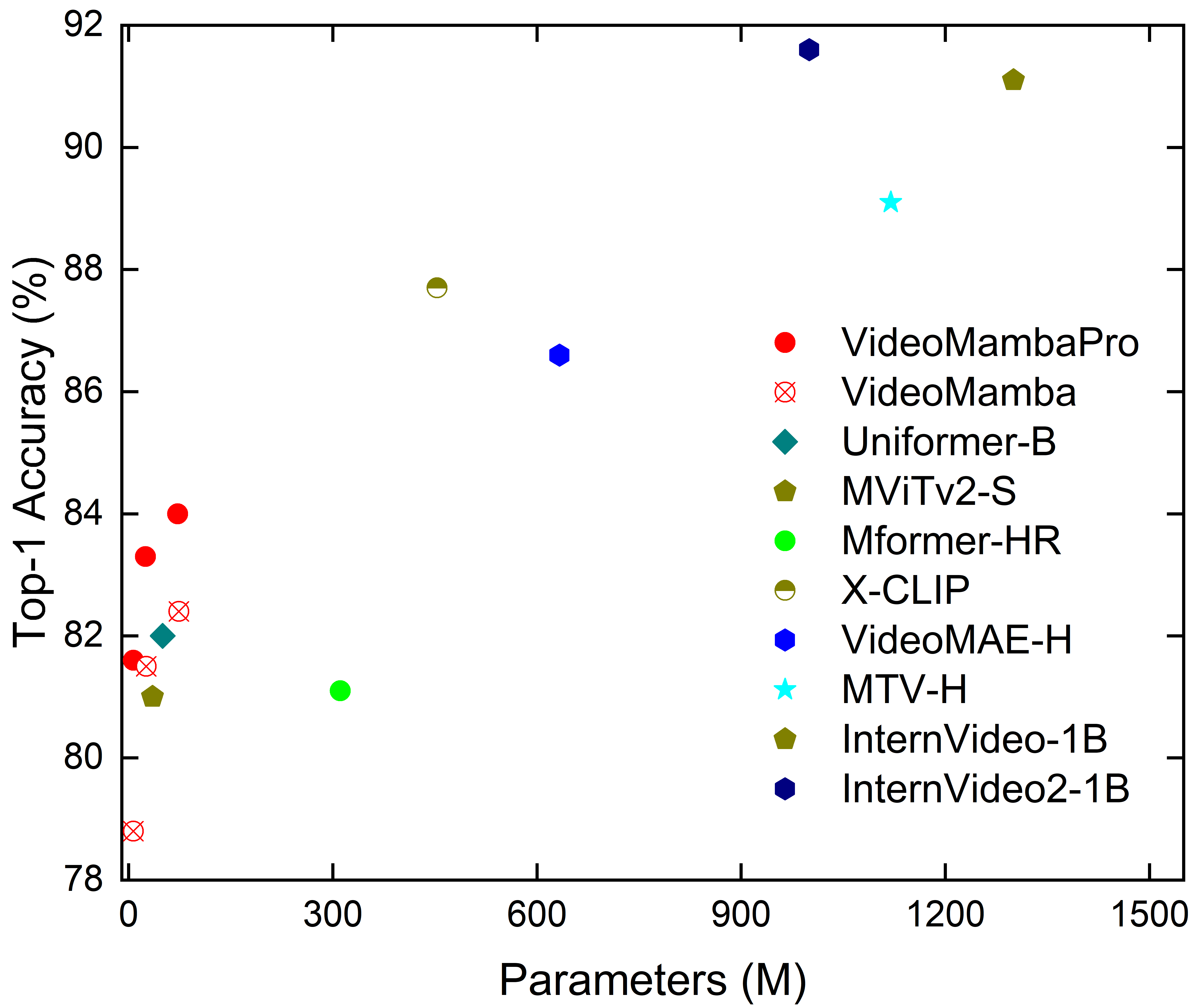}
   \caption{Top-1 accuracy versus number of parameters of VideoMambaPro and other models on Kinetics-400.}
   \label{fig:parameter comparison}
\end{figure}

\begin{figure}[htb]
  \centering
  \includegraphics[width=1.0\linewidth]{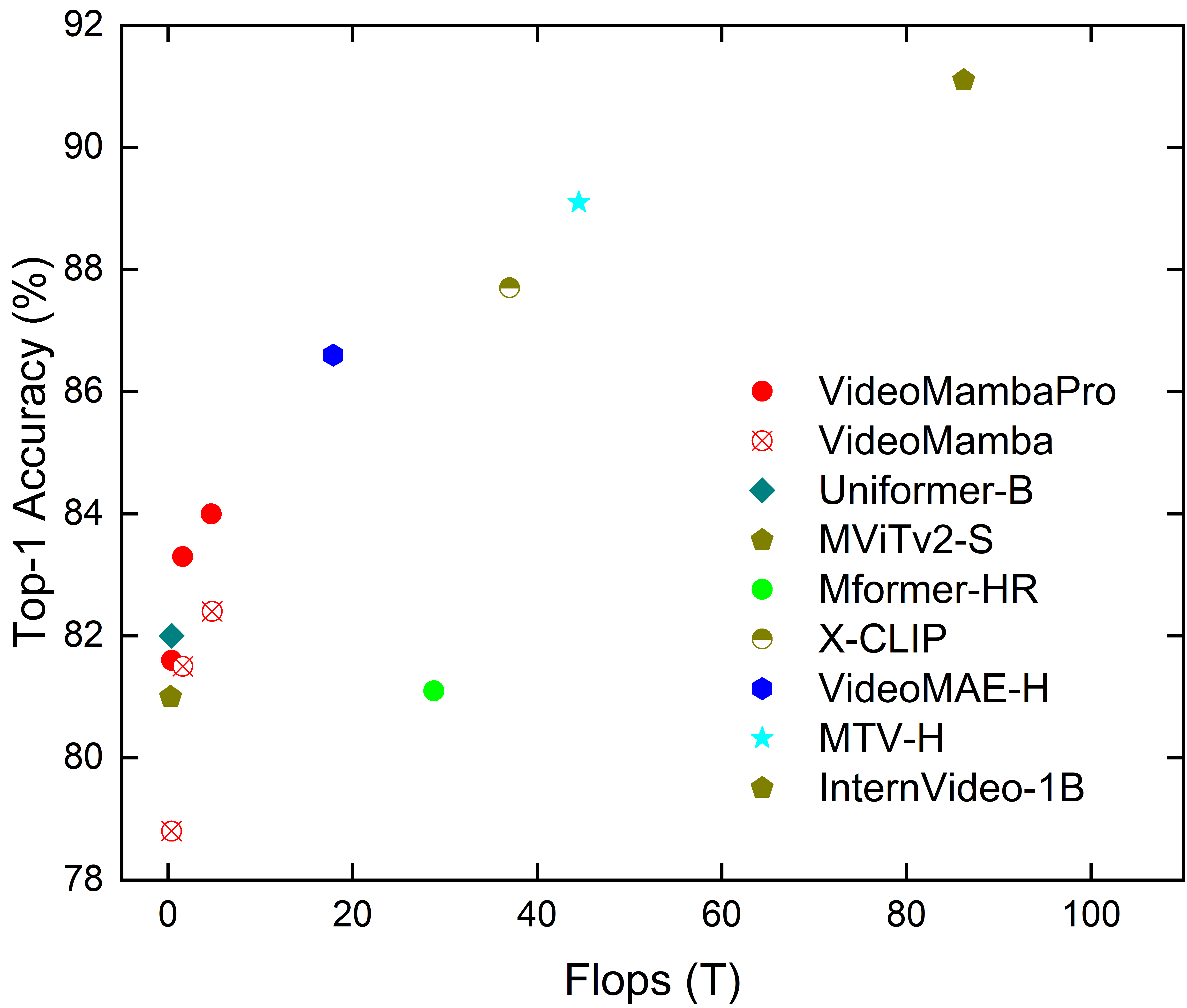}
   \caption{Top-1 accuracy versus number of FLOPs of VideoMambaPro and other models on Kinetics-400.}
   \label{fig:flop comparison}
\end{figure}

\subsection{Computation cost analysis}
\label{subsec:computation cost}
Finally, we compare the performance of VideoMambaPro with various model sizes with other approaches on K400. We map the top-1 to the number of parameters and FLOPs in Figures~\ref{fig:parameter comparison} and \ref{fig:flop comparison}, respectively. VideoMambaPro outperforms VideoMamba and transformers with similar parameter counts and FLOPs. Notably, as the parameter count and FLOPs increase, the performance improvement trend of our method is steeper than that of transformers. This suggests that with further enhancements to the Mamba architecture, such as introducing additional parameters, its performance could surpass current state-of-the-art models while requiring fewer parameters and FLOPs.

\section{Conclusion} \label{sec:conclusion}

From a mathematical comparison with self-attention, we have identified two limitations in how Mamba processes token sequences. We argue that these limitations constrain Mamba's potential, especially in video understanding tasks. To address the two limitations, we have introduced VideoMambaPro (VMP), which takes VideoMamba and introduces the masked backward State Space Model (SSM), and adds residual connections in both forward and backward SSM. In experiments on Kinetics-400, Something-Something V2, HMDB51, UCF-101, and AVA V2.2, VideoMambaPro consistently demonstrates improved performance over the vanilla VideoMamba. 
In this paper, we have refrained from extensive pre-training. We expect that this could further elevate the performance of Mamba models for video tasks, making it an increasingly attractive, efficient alternative to large transformer models.

{
    \small
    \bibliographystyle{ieeenat_fullname}
    \bibliography{main}
}

\newpage

We provide the architectures for VideoMambaPro models in Section~\ref{sec:appendix_architecture}. A comparison between the architectures of VideoMamba and VideoMambaPro appears in Section~\ref{sec:appendix_comparison}. Training details are shown in Section~\ref{sec:appendix_implementation}. Finally, we report ImageNet-1K image classification results in Section~\ref{sec:appendix_imagenet_1k}.

\section{VideoMambaPro architectures}
\label{sec:appendix_architecture}

We present the architecture details of VideoMambaPro-Tiny (Ti), -Small (S), and -Middle (M) in Tables~\ref{tab:archi_ti}--\ref{tab:archi_middle}. The differences are in the embedding dimension (192, 384, 576) and the number of SSM blocks (24, 24, 32).

\begin{table}[htb]
  \centering
  \resizebox{\linewidth}{!}{
  \begin{tabular}{cc}
    \toprule
    Stage & Tiny \\
    \midrule
    Patch Embedding & nn.Conv3d (kernel size = $16 \times 16 \times 1 $, \textbf{embedding dimension = 192})\\
    \midrule
    SSM & $\begin{array}{@{}c@{}}
    \begin{bmatrix}
    \text{MLP} (768) \\
    \text{MLP} (3072) \\
    \text{MHA (head = 12)}
    \end{bmatrix} \times \textbf{24}
    \end{array}$ \\
    \midrule
    \multirow{4}{*}{Projection} & Layer Normalization\\
    &Dropout (ratio)\\
    &Linear layer (1000)\\
    & Softmax\\
    \bottomrule \\
  \end{tabular}
  }
  \caption{Architecture details of VideoMambaPro-Ti.}
  \label{tab:archi_ti}
\end{table}

\begin{table}[htb]
  \centering
  \resizebox{\linewidth}{!}{
  \begin{tabular}{cc}
    \toprule
    Stage & 
    Small \\
    \midrule
    Patch Embedding & nn.Conv3d (kernel size = $16 \times 16 \times 1 $, \textbf{embedding dimension = 384})\\
    \midrule
    SSM & $\begin{array}{@{}c@{}}
    \begin{bmatrix}
    \text{MLP} (768) \\
    \text{MLP} (3072) \\
    \text{MHA (head = 12)}
    \end{bmatrix} \times \textbf{24}
    \end{array}$ \\
    \midrule
    \multirow{4}{*}{Projection} & Layer Normalization\\
    &Dropout (ratio)\\
    &Linear layer (1000)\\
    & Softmax\\
    \bottomrule \\
  \end{tabular}
  }
  \caption{Architecture details of VideoMambaPro-S.}
  \label{tab:archi_small}
\end{table}

\begin{table}[htb]
  \centering
  \resizebox{\linewidth}{!}{
  \begin{tabular}{cc}
    \toprule
    Stage & 
    Middle \\
    \midrule
    Patch Embedding & nn.Conv3d (kernel size = $16 \times 16 \times 1 $, \textbf{embedding dimension = 576})\\
    \midrule
    SSM & $\begin{array}{@{}c@{}}
    \begin{bmatrix}
    \text{MLP} (768) \\
    \text{MLP} (3072) \\
    \text{MHA (head = 12)}
    \end{bmatrix} \times \textbf{32}
    \end{array}$ \\
    \midrule
    \multirow{4}{*}{Projection} & Layer Normalization\\
    &Dropout (ratio)\\
    &Linear layer (1000)\\
    & Softmax\\
    \bottomrule \\
  \end{tabular}
  }
  \caption{Architecture details of VideoMambaPro-M.}
  \label{tab:archi_middle}
\end{table}

\section{Architecture comparison with VideoMamba}
\label{sec:appendix_comparison}
\begin{figure}[htb]
  \centering
  \includegraphics[width=0.9\linewidth]{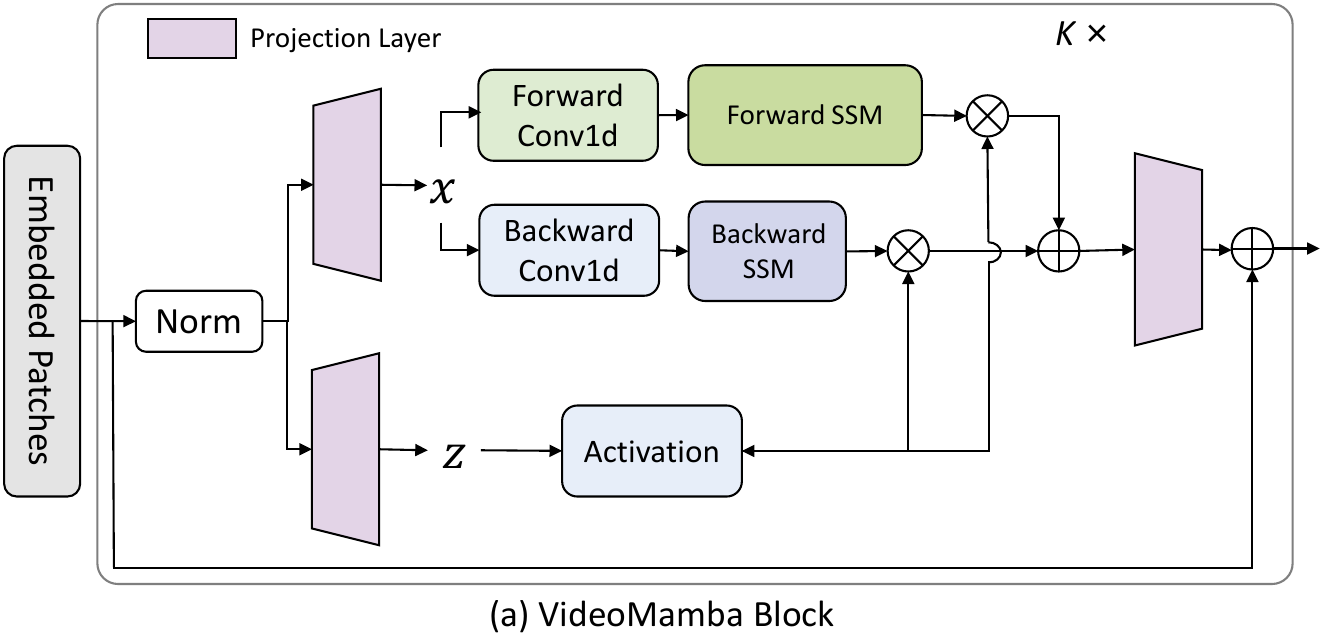}
  \includegraphics[width=0.9\linewidth]{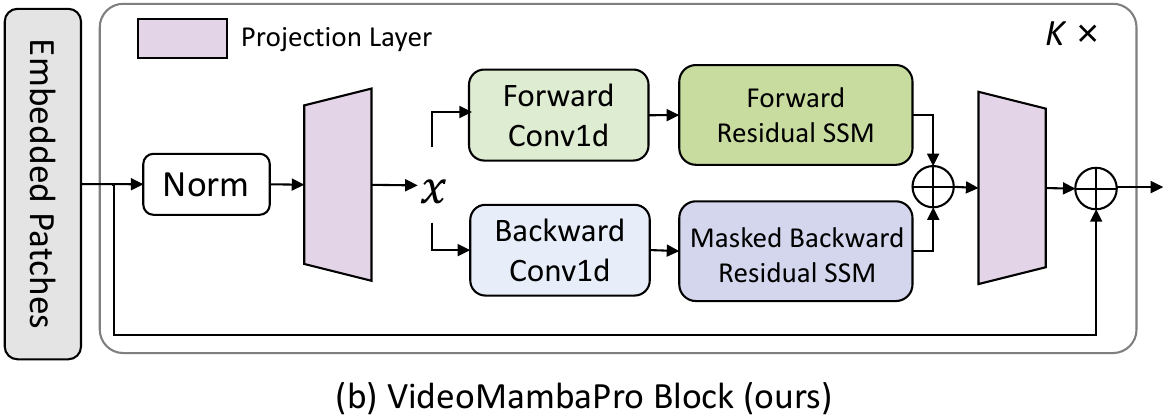}
   \caption{Comparison between the bi-directional VideoMamba (top) and VideoMambaPro (bottom) blocks.}
   \label{fig:archi comparison}
\end{figure}

We compare the architectures of VideoMambaPro and VideoMamba~\cite{li2024videomamba} in Figure~\ref{fig:archi comparison}. VideoMambaPro does not have the linear layer to generate parameters $z$. Additionally, our residual SSM and mask scheme do not introduce additional parameters or computational overhead, so our method has slightly fewer parameters and FLOPs.

\section{Implementation details}
\label{sec:appendix_implementation}
We conduct pre-training on ImageNet-1K and fine-tuning on the Something-Something V2 and Kinetics-400 datasets with 16 NVIDIA A100-80G GPUs. Models for UCF101 and HMDB51 are trained with 8 A100-80G GPUs. The experiments on AVA are conducted with 32 A100-80G GPUs. The values of the hyperparameters are largely similar to those used in VideoMamba~\cite{li2024videomamba}.
We linearly scale the base learning rate with respect to the overall batch size, $lr = lr_{base} \times batch size / 256$. The pre-training details are shown in Table~\ref{tab:ImageNet-1K}, and the fine-tuning details on the other datasets are listed in Tables~\ref{tab:k400}--\ref{tab:ava}.

\begin{table}[htb]
  \centering
  \resizebox{0.8\linewidth}{!}{
  \begin{tabular}{lc}
    \toprule
    config &image size: $224 \times 224$ \\
    \midrule
    optimizer & AdamW\\
    base learning rate & 1.5e-4\\
    weight decay & 0.1 (Tiny), 0.05 (Small, Middle)\\
    minimal learning rate & 1.0e-6 \\
    optimizer momentum& $\beta_1, \beta_2$ = 0.9, 0.95\\
    batch size & 512 \\
    learning rate schedule & cosine decay \\
    warmup epochs & 5 (Tiny), 10 (Small), 40 (Middle)\\
    dropout ratio & 0 (Tiny), 0.15 (Small), 0.5 (Middle)\\
    augmentation & MultiScaleCrop \\
    label smoothing & 0.1\\
    \bottomrule
  \end{tabular}
  }
  \caption{Pre-training setting on ImageNet-1K}
  \label{tab:ImageNet-1K}
\end{table}

\begin{table}[htb]
  \centering
  \resizebox{\linewidth}{!}{
  \begin{tabular}{lc}
    \toprule
    config & image size: $224 \times 224$ \\
    \midrule
    optimizer & AdamW\\
    base learning rate & 1.5e-4\\
    weight decay & 0.1 (Tiny), 0.05 (Small, Middle)\\
    minimal learning rate & 1.0e-6 \\
    optimizer momentum& $\beta_1, \beta_2$ = 0.9, 0.99\\
    batch size & 256 \\
    learning rate schedule & cosine decay \\
    warmup epochs & 5 (Tiny), 5 (Small) 10 (Middle)\\
    dropout ratio & 0.1 (Tiny), 0.35 (Small), 0.6 (Middle)\\
    augmentation & RandAug (7, 0.25) (Tiny), RandAug (9, 0.5) (Small, Middle) \\
    label smoothing & 0.1\\
    flip augmentation & yes\\
    \bottomrule
  \end{tabular}
  }
  \caption{Fine-tuning setting for Kinetics-400}
  \label{tab:k400}
\end{table}

\begin{table}[htb]
  \centering
  \resizebox{\linewidth}{!}{
  \begin{tabular}{lc}
    \toprule
    config & image size: $224 \times 224$ \\
    \midrule
    optimizer & AdamW\\
    base learning rate & 4e-4\\
    weight decay & 0.1 (Tiny), 0.05 (Small, Middle)\\
    minimal learning rate & 1.0e-6 \\
    optimizer momentum& $\beta_1, \beta_2$ = 0.9, 0.999\\
    batch size & 256 \\
    learning rate schedule & cosine decay \\
    warmup epochs & 5 (Tiny), 5 (Small) 10 (Middle)\\
    dropout ratio & 0.1 (Tiny), 0.35 (Small), 0.6 (Middle)\\
    augmentation & RandAug (7, 0.25) (Tiny), RandAug (9, 0.5) (Small, Middle)\\
    label smoothing & 0.1\\
    flip augmentation & no\\
    \bottomrule
  \end{tabular}
  }
  \caption{Fine-tuning setting for Something-Something V2}
  \label{tab:ssv2_finetuning}
\end{table}

\begin{table}[htb]
  \centering
  \resizebox{\linewidth}{!}{
  \begin{tabular}{lc}
    \toprule
    config & image size: $224 \times 224$ \\
    \midrule
    optimizer & AdamW\\
    base learning rate & 4e-4\\
    weight decay & 0.1 (Tiny), 0.05 (Small, Middle)\\
    minimal learning rate & 1.0e-6 \\
    optimizer momentum& $\beta_1, \beta_2$ = 0.9, 0.99\\
    batch size & 128 \\
    learning rate schedule & cosine decay \\
    warmup epochs & 5 (Tiny), 5 (Small) 10 (Middle)\\
    dropout ratio & 0.1 (Tiny), 0.35 (Small), 0.6 (Middle)\\
    augmentation & RandAug (7, 0.25) (Tiny), RandAug (9, 0.5) (Small, Middle)\\
    label smoothing & 0.1\\
    flip augmentation & yes\\
    \bottomrule
  \end{tabular}
  }
  \caption{Fine-tuning setting for UCF101/HMDB51}
  \label{tab:ucf and hmdb}
\end{table}

\begin{table}[htb]
  \centering
  \resizebox{\linewidth}{!}{
  \begin{tabular}{lc}
    \toprule
    config & image size: $224 \times 224$ \\
    \midrule
    optimizer & AdamW\\
    base learning rate & 1.5e-3 (Tiny), 2.5e-4 (Small, Middle) \\
    weight decay & 0.051 (Tiny, Small, Middle)\\
    minimal learning rate & 1.0e-6 \\
    optimizer momentum& $\beta_1, \beta_2$ = 0.9, 0.999\\
    batch size & 128 \\
    learning rate schedule & cosine decay \\
    warmup epochs & 5 (Tiny), 5 (Small) 10 (Middle)\\
    dropout ratio & 0.1 (Tiny), 0.35 (Small) 0.6 (Middle)\\
    augmentation & RandAug (7, 0.25) (Tiny), RandAug (9, 0.5) (Small, Middle)\\
    label smoothing & 0.1\\
    flip augmentation & yes\\
    \bottomrule
  \end{tabular}
  }
  \caption{Fine-tuning setting for AVA 2.2}
  \label{tab:ava}
\end{table}

\section{Results on ImageNet-1K} \label{sec:appendix_imagenet_1k}
Before moving to the video domain, we pre-train VideoMambaPro on ImageNet-1K, which contains 1.28M training images and 50K validation images across 1,000 categories. All models are trained on the training set, and top-1 accuracy on the validation set is reported. For fair comparison, we adopt the same method as VideoMamba, and our training settings primarily follow DeiT~\cite{touvron2021training}. When training on $224 ^ 2$ input images, we use AdamW with a momentum of 0.9 and a total batch size of 512. Training is performed on 8 A800 GPUs, with more details provided in Table~\ref{tab:ImageNet-1K}. The results are summarized in Table~\ref{tab:imagenet results}. VideoMambaPro achieves accuracy gains of 1.0-2.0\% over VideoMamba. 

\begin{table}[htbp]
\resizebox{\linewidth}{!}{
\begin{tabular}{lcccc}
\toprule
                    & Input & Param & FLOPs & Top-1 \\ \midrule
VideoMamba (Ti)    & 224 $^2$  & 7M        & 1.1G      & 76.9  \\ 
VideoMambaPro (Ti) & 224 $^2$  & 7M        & 1.1G      & 78.9  \\ \hline
VideoMamba (S)    & 224 $^2$  & 26M       & 4.3G      & 81.2  \\ 
VideoMambaPro (S) & 224 $^2$  & 25M       & 4.2G      & 82.4  \\ \hline
VideoMamba (M)    & 224 $^2$  & 74M       & 12.7G     & 82.8  \\ 
VideoMambaPro (M) & 224 $^2$  & 72M       & 12.4G     & 83.8  \\ \bottomrule
\end{tabular}
}
\caption{ImageNet-1K pre-training results for VideoMamba and VideoMambaPro.}
  \label{tab:imagenet results}
\end{table}

\end{document}